\documentclass{article}

\usepackage{microtype}
\usepackage{graphicx}
\usepackage{booktabs} 
\usepackage{amssymb}
\usepackage{enumerate}
\usepackage{subcaption}
\usepackage{amsmath}

\usepackage{hyperref}

\usepackage[accepted]{icml2019}

\icmltitlerunning{Robust Anomaly Detection using Adversarial Autoencoders}

\begin{document}

\twocolumn[
\icmltitle{Robust Anomaly Detection in Images using Adversarial Autoencoders}



\icmlsetsymbol{equal}{*}

\begin{icmlauthorlist}
\icmlauthor{Laura Beggel}{rb,lmu}
\icmlauthor{Michael Pfeiffer}{rb}
\icmlauthor{Bernd Bischl}{lmu}
\end{icmlauthorlist}

\icmlaffiliation{lmu}{Department of Statistics, Ludwig-Maximilians-University Munich, Munich, Germany}
\icmlaffiliation{rb}{Bosch Center for Artificial Intelligence, Renningen, Germany}

\icmlcorrespondingauthor{Laura Beggel}{laura.beggel@googlemail.com}

\icmlkeywords{Machine Learning, ICML}

\vskip 0.3in
]



\printAffiliationsAndNotice{}  

\begin{abstract}
Reliably detecting anomalies in a given set of images is a task of high practical relevance for visual quality inspection, surveillance, or medical image analysis. 
Autoencoder neural networks learn to reconstruct normal images, and hence can classify those images as anomalies, where the reconstruction error exceeds some threshold. 
Here we analyze a fundamental problem of this approach when the training set is contaminated with a small fraction of outliers.
We find that continued training of autoencoders inevitably reduces the reconstruction error of outliers, and hence degrades the anomaly detection performance. 
In order to counteract this effect, an adversarial autoencoder architecture is adapted, which imposes a prior distribution on the latent representation, typically placing anomalies into low likelihood-regions.
Utilizing the likelihood model, potential anomalies can be identified and rejected already during training, which results in an anomaly detector that is significantly more robust to the presence of outliers during training.
\end{abstract}

\section{Introduction}
\label{sec:introduction}
The goal of anomaly detection is to identify observations in a dataset that significantly deviate from the remaining observations \cite{hawkins1980identification}. 
Since anomalies are rare and of diverse nature, it is not feasible to obtain a labeled dataset representative of all possible anomalies. A successful approach for anomaly detection is to learn a model of the normal class, under the assumption that the training data consists entirely of normal observations. If an observation deviates from that learned model, it is classified as an anomaly \cite{chandola2009anomaly}.
Autoencoder neural networks have shown superior performance for anomaly detection on high dimensional data such as images. 
Autoencoders consist of an encoder network, which performs nonlinear dimensionality reduction from the input into a lower-dimensional latent representation, followed by a decoder network, which reconstructs the original image from the latent representation. 
Autoencoders do not require label information since the input image also represents the desired output. By learning to extract features and to reconstruct the original images, the network yields a model that generalizes to the reconstruction of images similar to those in the training set. Conversely, images which show significant deviations from those observed during training will lead to reconstruction errors. The reconstruction error of an image can thus be used as an anomaly score. 

Although the autoencoder approach performs well on benchmark datasets \cite{williams2002comparative}, we identify in this article several major shortcomings for real-world scenarios.
First, autoencoder methods for anomaly detection are based on the assumption that the training data  consists only of instances that were previously confirmed to be normal. In practice, however, a clean dataset cannot always be guaranteed, e.g., because of annotation errors, or because inspection of large datasets by domain experts is too expensive or too time consuming. It is therefore desirable to learn a model for anomaly detection from completely unlabeled data, thereby risking that the training set is contaminated with a small proportion of anomalies. 
However, we find that autoencoder-based anomaly detection methods are very sensitive to even slight violations of the clean-dataset assumption. A small number of anomalies contaminating the training might result in the autoencoder learning to reconstruct  anomalous observations as well as  normal ones. We analyze the underlying causes for this vulnerability of standard autoencoders, and present several key ideas that make anomaly detection with autoencoders more robust to training anomalies, thereby improving the overall anomaly detection performance.

In summary, our contributions are:
First, we use adversarial autoencoders \cite{makhzani2015adversarial}, which allow to control the distribution of latent representations, thereby defining a prior distribution in the bottleneck layer. While (adversarial) autoencoders have been used for anomaly detection before \cite{zhai2016deep, leveau2017adversarial}, we here propose a novel combined criterion of reconstruction error and likelihood in latent space. Since anomalies are expected to have a low likelihood under the given prior of the normal data, the combination of likelihood and reconstruction error yields an improved anomaly score and therefore better detection performance. 
Second, we define an iteration refinement method for training sample rejection. Potential anomalies in the training set are identified in the lower dimensional latent space by a variation of 1-class SVM \cite{scholkopf2001estimating}, and by rejecting the least normal observations we can increase robustness to contaminated data.
Third, we propose a retraining method to increase separation in both latent and image space.
We compare our method to \cite{hinton2006reducing, makhzani2015adversarial}, and show that our proposed combination results in an anomaly detector with significantly increased robustness against anomalies present during training.

\section{Related work}
Autoencoders were originally intended for nonlinear dimensionality reduction and feature extraction \cite{hinton2006reducing}, but it has been realized early on that their capability to model the training data distribution makes them suitable for anomaly detection \cite{japkowicz1995novelty}.
More recent work has proposed probabilistic interpretations of deep autoencoders, which can directly model aspects of the data generating process.
Denoising autoencoders \cite{vincent2008extracting} learn reconstruction of images from noise corrupted inputs. This form of regularization makes the latent representation focus on a data manifold which encodes the most relevant image features. In \cite{alain2014regularized, bengio2013generalized} it was shown that regularized autoencoders implicitly estimate the data generating process, and have established links between reconstruction error and the data generating density. \cite{zhai2016deep} applied these concepts to anomaly detection with deep structured energy based models, showing that a criterion based on an energy score leads to better results than the reconstruction error criterion.
Adversarial Autoencoders (AAE) \cite{makhzani2015adversarial} learn a generative model of the input data by combining the reconstruction error with an adversarial training criterion \cite{goodfellow2014generative}. A discriminator network learns to distinguish between samples coming from the encoder and from a desired arbitrary prior distribution, which gives AAEs great flexibility to represent assumptions about the data distribution. 
AAEs for anomaly detection were first proposed in \cite{leveau2017adversarial}, using a Gaussian mixture model as prior. It was found that a purely unsupervised approach did not separate anomalies and normal images into different clusters, and it was proposed to either condition on class labels, or train an explicit rejection class with random images.

Almost all approaches for anomaly detection with autoencoders require the training data to consist of normal examples only, but this alone is no guarantee for anomalies to have large reconstruction errors.
Robust deep autoencoders \cite{zhou2017anomaly} address this issue by combining denoising autoencoders with robust PCA, thereby isolating noise and outliers from training of the reconstruction. The method achieves significantly better results in the presence of anomalies in the training set on MNIST.
A combination of deep learning and kernel based methods for anomaly detection in high dimensional data was proposed by \cite{erfani2016high}, who combine a Deep Belief Network for feature extraction, and a 1-class SVM for anomaly detection in the compressed latent space. Their method can deal with anomalies in the training data, but does not use this information to refine the training set.
When also considering adversarial detection, \cite{feinman2017detecting} have proposed density based measures in a ConvNet to identify data points that lie outside the data manifold as potential adversarial examples. They increase the robustness of their method by adding a Bayesian uncertainty estimate, which handles complementary situations.

\section{Autoencoders and their Limitations}
\label{sec:ae}

\begin{figure*}[ht]
	\centering
	\begin{subfigure}{\columnwidth}
		\resizebox{.9\linewidth}{!}{ 
			\includegraphics[trim=5 5 0 0,clip,width=\columnwidth] {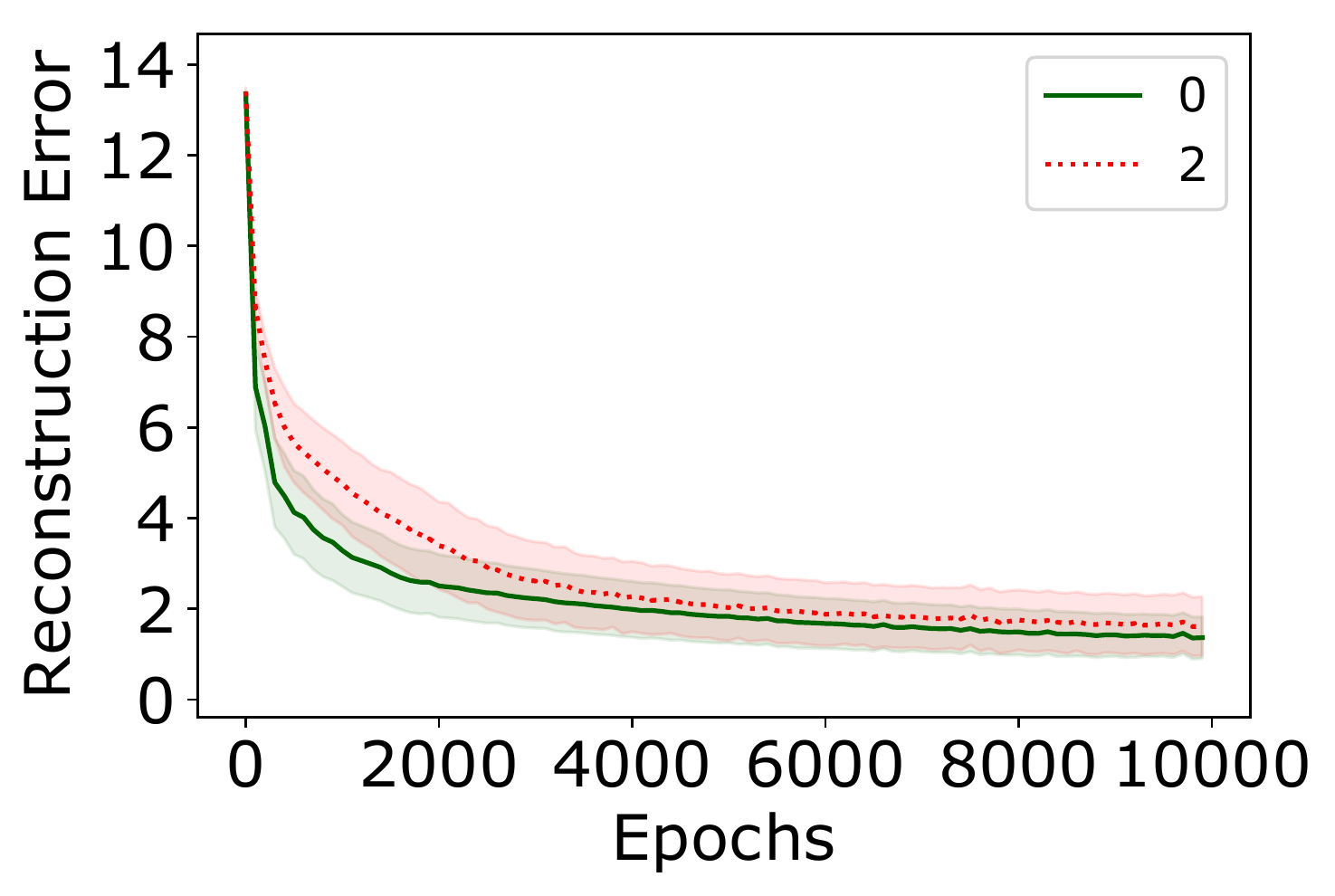} 
		}
		\caption{} 
	\end{subfigure} \hfill
	\begin{subfigure}{\columnwidth}
		\resizebox{.9\linewidth}{!}{
		\includegraphics[width=\columnwidth]{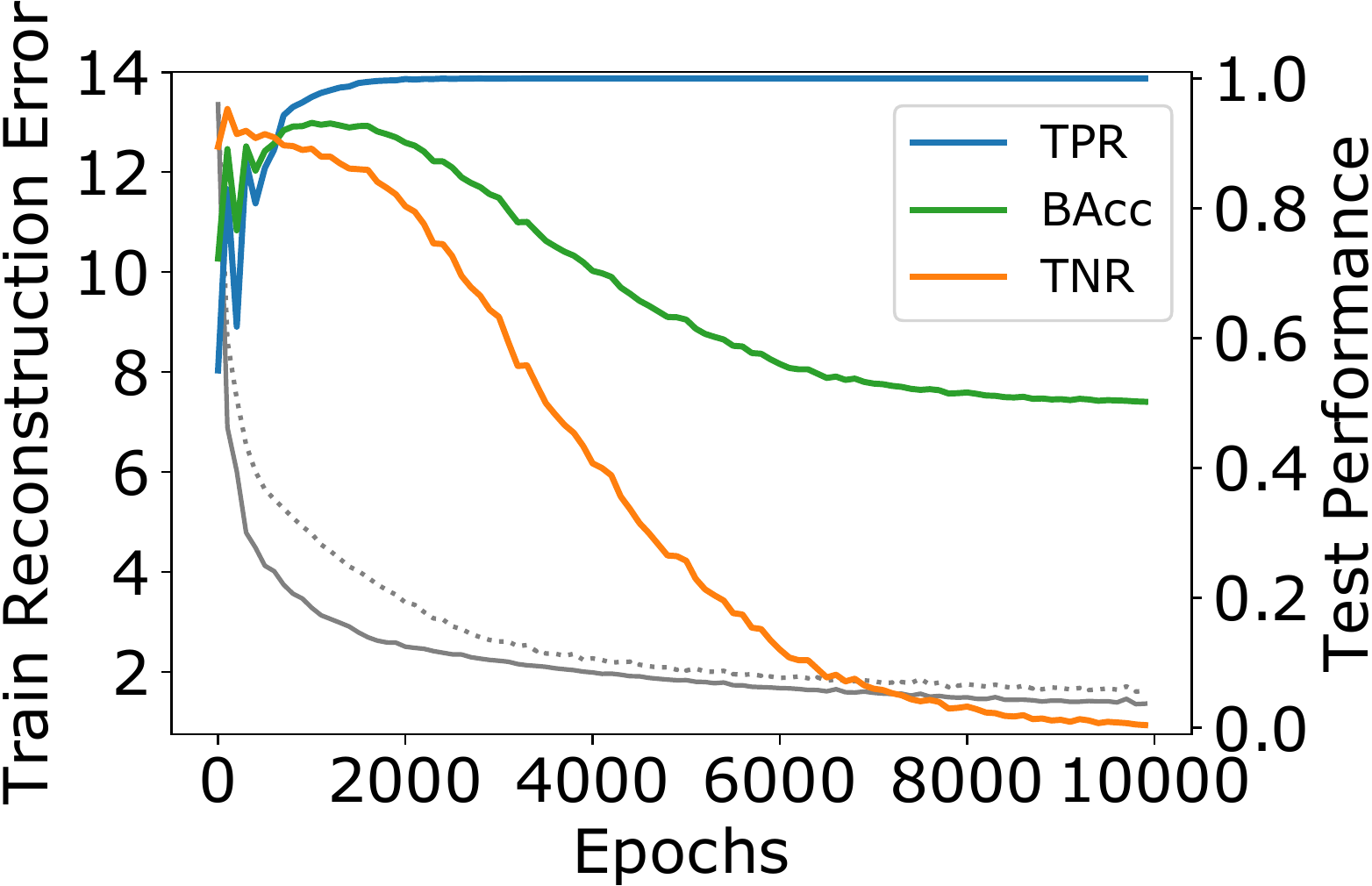} 
		}
		\caption{}
	\end{subfigure}
	\caption{
		Limitations of conventional autoencoders for anomaly detection. 
		(a) Mean reconstruction error of a conventional AE trained on  MNIST, where $95\%$ of the images are from the normal class (digit '0', green, solid line), and $5\%$ are anomalies (digit '2', red, dashed line). The shaded area shows the standard deviation. As training progresses, the AE learns to reconstruct the anomalous as well as the normal images. (b) Detection performance on test data from MNIST, measured by the True Positive Rate (TPR), True Negative Rate (TNR), and Balanced Accuracy (BAcc), where the reconstruction threshold is set to the $90$th percentile. The gray lines indicate the mean training reconstruction error as displayed in (a). As training progresses, the AE produces more and more false positives, since the distribution of reconstruction errors between normal and anomalous images increasingly overlap.}
	\label{fig:ae}
\end{figure*}

An autoencoder (AE) is a neural network that maps an input image $\mathbf{x} \in \mathcal{X} = \mathbb{R}^n$ to an output image $\mathbf{x}^\prime \in \mathcal{X}$. It consists of an encoder function $f: \mathcal{X} \rightarrow \mathcal{Z}$ and a decoder function $g: \mathcal{Z} \rightarrow \mathcal{X}$, each implemented as a multi-layer neural network. They jointly compute $\mathbf{x}^\prime = g(f(\mathbf{x}))$. The output of the encoder $\mathbf{z} = f(\mathbf{x}) \in \mathcal{Z} = \mathbb{R}^m \ (m \ll n)$ is a low-dimensional latent representation of $\mathbf{x}$. This bottleneck prevents the AE from learning a trivial identity function. The autoencoder is trained to minimize the reconstruction error $L(\mathbf{x}, \mathbf{x}^\prime)$, which is typically the pixelwise mean squared error or the Euclidean distance in the image space $\mathcal{X}$. After training, anomaly detection can be performed by comparing $L(\mathbf{x}, \mathbf{x}^\prime)$ to a decision threshold $T_{\mathrm{rec}}$, classifying all images $\mathbf{y}$ with $L(\mathbf{y}, g(f(\mathbf{y})))>T_{\mathrm{rec}}$ as anomalies. $T_{\mathrm{rec}}$ is selected based on the distribution of all reconstruction errors $L_{\mathrm{train}}$ on the training set $\mathbf{X}_{\mathrm{train}}$. Typical choices are the maximum reconstruction error $T_{\mathrm{rec}} = \max_{\mathbf{x}\in\mathbf{X}_{\mathrm{train}}} L(\mathbf{x}, \mathbf{x}^\prime)$, or a large percentile (e.g., 90\%) $T_{rec} = p_{0.9}(L(\mathbf{x}, \mathbf{x}^\prime) |\mathbf{x}\in\mathbf{X}_{\mathrm{train}})$, which is more robust. 
Using autoencoders for detecting anomalies with this procedure is based on the assumption that all training examples should be reconstructed well, or in other words that the training set is clean and consists only of normal observations.

\paragraph{Training with anomalies}
A standard autoencoder learns to reconstruct images from an intrinsic lower dimensional latent representation, and by simultaneously learning a mapping from image into latent space also learns in its weights an implicit model of the data it has seen.
For the task of anomaly detection this leads to a trade-off between generating reconstructions of previously unseen normal images with minimal error, while maximizing the reconstruction error of anomalous images. Since no labels are available during training, neither of the criteria can be directly optimized. Instead, the AE is trained to minimize reconstruction errors on the entire training set, which will only directly optimize the first criterion if all training images are normal. 
During training, the objective rewards exact reconstructions of all training images, including anomalies.
Overfitting singular anomalies can be avoided by reducing model capacity or early stopping, such that the AE focuses on reconstructing the majority class. 
Early stopping, however, may prevent the autoencoder from learning a model which can precisely reconstruct the majority of (normal) training observations, and may thus lead to false detections.

We demonstrate this effect for a conventional autoencoder trained on two classes of images of handwritten digits from MNIST \cite{lecun1998gradient}. A detailed description of the architecture can be found in Sec.~\ref{sec:experiments}.
The digit '0' is arbitrarily defined as the normal class, whereas digit '2' is the anomaly class (different combinations of digits lead to similar results). In this experiment the training set includes $5\%$ anomalies. 
Fig.~\ref{fig:ae}(a) shows the reconstruction error for a conventional AE trained over $10000$ epochs, which results in a network that reconstructs both classes with very similar error.
Using early stopping as proposed in \cite{zhou2017anomaly, vincent2010stacked}, e.g., after  only $100$ or $1000$ iterations results in a model that is better at reconstructing normal compared to anomalous images, but it has not yet learned an accurate reconstruction model for the normal class.
Convergence is reached only after more than $4000$ epochs, but at that time the model reconstructs both normal and anomalous images equally well. This results in poor performance as an anomaly detector, as shown in Fig.~\ref{fig:ae}(b).

We evaluate the True Positive Rate (TPR), True Negative Rate (TNR), and Balanced Accuracy (BAcc) at different epochs (where an anomaly is a positive event). BAcc is defined as $\frac{\mathrm{TPR} + \mathrm{TNR}}{2} \in [0,1]$ and thus balances detection performance \cite{brodersen2010balanced}. 
We do not use the F1 score, which is commonly used in anomaly detection, since it neglects the true negative prediction performance.
Clearly, the importance of each metric depends on the role that false negatives (i.e., missed anomalies) and false alarms have for the task at hand. 
But obviously, approaching a TPR of 1 at the cost of a TNR going towards 0 (as is the case for an autoencoder trained until convergence) is not desirable. 
For the evaluation we use the known labels of images, which are however never used during training.

An immediate observation from Fig.~\ref{fig:ae}(b) is that continued training leads to a drop in TNR and thus BAcc, which is due to increasing overlap between the distribution of reconstruction errors of normal and anomalous images.
A possible explanation for this behavior lies in the nature of stochastic gradient descent, the method used to train autoencoders. In the initial phase, the AE learns to reconstruct the normal class, which is heavily overrepresented in the training set, and thus leads to more training updates. This effect is visible in Fig.~\ref{fig:ae}(a), where the reconstruction error of the normal class shrinks much faster initially than that of anomalous examples. After a few hundred epochs, the error for normal images continues to shrink slowly, but the error for anomalies falls faster. 
This is due to the small gradient for normal examples, whereas anomalies with still large errors result in large gradients, and therefore dominate the direction of updates. As a result, the difference in reconstruction quality between normal and anomalous images vanishes at later epochs.
One strategy could be to reduce model capacity, with the hope that in a smaller network only the majority class can be accurately reconstructed. However, this strategy also results in lower quality reconstructions for normal images, and therefore in a higher reconstruction threshold, which is again prone to yielding many false negatives. A similar argument explains why early stopping does not solve the issue.

\begin{figure}[ht]
	\centering	
	\resizebox{.9\linewidth}{!}{
			\includegraphics[width=0.9\columnwidth]{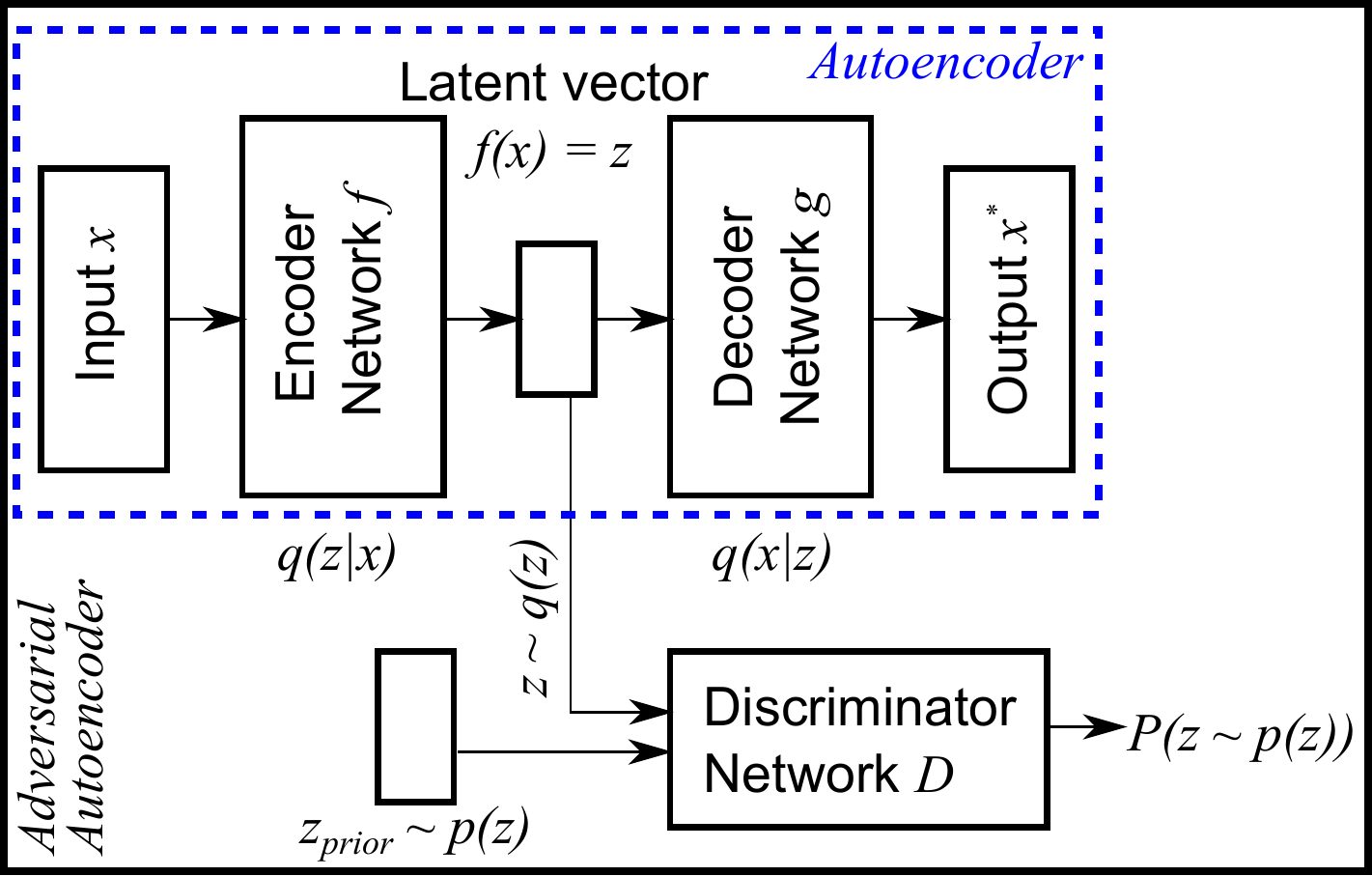} 		
		}
		\caption{Schematic structure of conventional autoencoder (blue dashed box) and the extension to an adversarial autoencoder. }
		\label{fig:Struct}
	\end{figure}

\paragraph{Adversarial Autoencoders} 
\label{sec:aae} 
Adversarial autoencoders (AAE) \cite{makhzani2015adversarial} extend the concept of autoencoders by inducing a prior distribution $p(\mathbf{z})$ in the latent space. A generative model of the data distribution $p_{\mathrm{data}}(\mathbf{x})$ is thus obtained by applying the decoder to samples from the imposed prior in latent space.
The main difference to Variational autoencoders \cite{kingma2014auto} is the use of an adversarial training criterion \cite{goodfellow2014generative}. As a result, AAEs can impose any prior distribution from which samples can be drawn, and have smoothly varying outputs in data space for small changes in corresponding latent space. An example of an AAE structure is displayed in Fig.~\ref{fig:Struct}.

\begin{figure}![ht]
	\centering	
	\resizebox{.9\linewidth}{!}{
	\includegraphics[trim=7 10 7 10,clip,width=0.9\columnwidth]{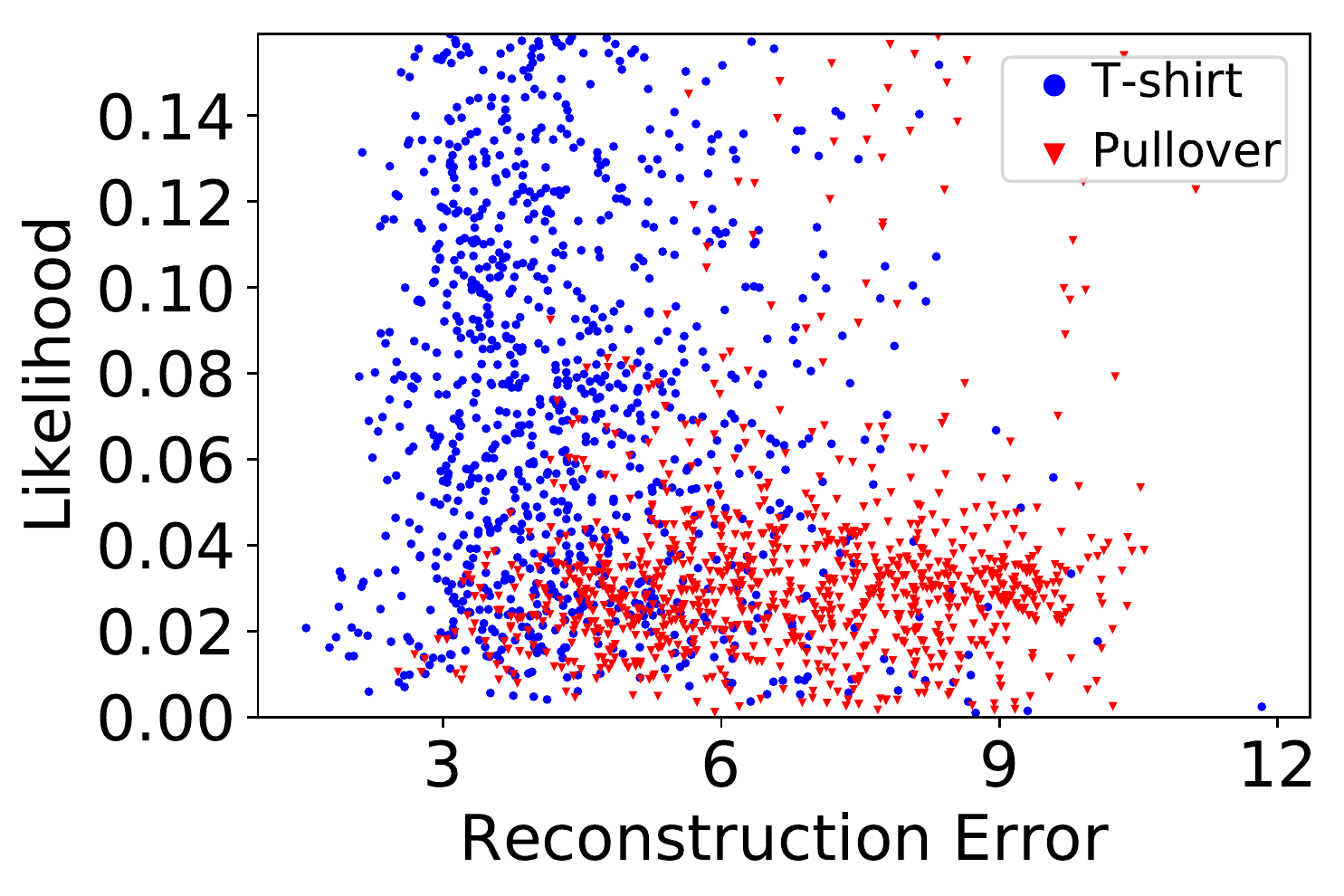} 
	}
	\caption{Reconstruction error and likelihood for an AAE trained on a clean subset of Fashion-MNIST containing only class 'T-shirt' (blue). Test data from the anomaly class 'Pullover' (red) yield lower likelihood values and higher reconstruction errors.}
	\label{fig:Example}
\end{figure}

From the perspective of anomaly detection AAEs are interesting because apart from the reconstruction error, the latent code provides an additional indication for anomalies \cite{leveau2017adversarial}. Simply put, we expect anomalies $\mathbf{x}$
(characterized by low $p_{\mathrm{data}}(\mathbf{x})$) to map to latent representations with low density $p(\mathbf{z}|\mathbf{x})$, or otherwise  have high reconstruction error $L(\mathbf{x}, \mathbf{x}^\prime)$, because high likelihood latent codes should be decoded into normal images (see Fig.~\ref{fig:Example}).

The previous analysis suggests a strategy to improve the robustness of autoencoders for anomaly detection in the presence of anomalies in the training set: If anomalies in the training set can be identified during training, there are ways to prevent a further improvement of their reconstruction quality. The simplest such strategy is to remove them from the training set, but other options are possible.
In the following we present a novel mechanism based on AAE, which actively manipulates the training set during training by sample rejection, and thereby focuses AE training on the normal class.

\section{Robust Anomaly Detection}
If the training set contains anomalies, then the AAE will model them as part of its generative model for $p_{\mathrm{data}}$, leading in principle to the same fundamental problem encountered in standard AE. 
However, depending on the imposed prior, we can at least expect a separation in latent space between the normal and anomalous instance encodings, since AAEs have smoothly varying outputs for nearby points in latent space. This feature of AAEs can be explicitly utilized by defining a prior distribution with a dedicated rejection class for anomalies \cite{leveau2017adversarial}, but we have observed the same effect even in the case of unimodal priors such as Gaussians.

Separation between anomalies and normal instances in latent space is particularly useful if a rough estimate of the training anomaly rate $\alpha$ is known. In this case standard outlier detection methods such as 1-class SVM \cite{scholkopf2001estimating} can be employed on the latent representations, searching for a boundary that contains a fraction of $1-\alpha$ of the whole dataset. Once potential anomalies are identified, they can be excluded for further training, 
or their contribution to the total loss might be reweighted. Such procedure approximates the case of a clean training set, where the combination of reconstruction error and latent density yields reliable results.


\subsection{Likelihood-based Anomaly Detection}
\label{sec:likelihood}
Since AAEs impose a prior distribution $p(\mathbf{z})$ on the latent representations $\mathbf{z}$, the likelihood $p(\hat{\mathbf{z}})$ under the prior of a new code vector $\hat{\mathbf{z}} = f(\hat{\mathbf{x}})$ can be used as an anomaly score \cite{leveau2017adversarial}. Anomalies are expected to have lower scores than normal examples. However, it is also clear that $p(\hat{\mathbf{z}})$ alone is an imperfect score, because anomalies in local clusters with small support might indeed be assigned higher scores than normal examples in boundary regions. Furthermore, the encoder might not be able to learn a mapping that exactly reproduces the prior. Despite these weaknesses, a likelihood-based criterion is able to identify most anomalies with similar performance as a reconstruction-based approach, and in addition allows a combination of both approaches.

A decision threshold $T_{\mathrm{prior}}$ is defined by measuring the likelihood $p(f(\mathbf{x}))$ under the imposed prior for all training samples $\mathbf{x}$ and then selecting a specified percentile in the distribution of $p(f(\mathbf{x}))$ depending on the expected anomaly rate $\alpha$. New examples $\mathbf{y}$ with $p(f(\mathbf{y}))<T_{\mathrm{prior}}$ are then classified as anomalies. Ideally we could set $T_{\mathrm{prior}}$ to the $\alpha$ percentile, but in practice the criterion is chosen slightly differently to compensate for approximation errors in the encoder and for biases induced by a finite training set. 
In our scenarios, $p_{0.1}(f(\mathbf{x}))$ was chosen empirically as it showed most robust behavior throughout all experiments. 
In the case of a clean dataset one can also fix the threshold, e.g., to a specified number of standard deviations, without optimizing on the training set.
Likelihood-based anomaly detection can be easily combined with reconstruction-based methods, and our results have shown that they complement each other well. We choose a simple combination whereby a new example $\mathbf{y}$ is classified as an anomaly if either $L(\mathbf{y}, \mathbf{y}^\prime) > T_{\mathrm{rec}}$, or $p(f(\mathbf{y})) < T_{\mathrm{prior}}$. Alternative methods such as a 1-class SVM in the 2-dimensional space of reconstruction errors and likelihoods did not improve our results. Although we focus on these two measures, it is also straightforward to integrate more criteria, such as denoising performance \cite{vincent2008extracting}, or sensitivity-based measures \cite{chan2017sensitivity}.

To compare the individual performance to a combination of both measures, we trained an AAE on a clean dataset consisting only of 'T-shirt's from Fashion-MNIST \cite{xiao2017online} (cf. Fig.~\ref{fig:Example}). For new test observations stemming from the normal class and a previously unseen anomaly class ('Pullover'), both the reconstruction error and the likelihood estimate identify anomalies with similar performance (BAcc: $0.72$ and $0.73$, respectively), and a combination of both criteria increases performance (BAcc: $0.80$). The architecture is described in detail in Sec.~\ref{sec:experiments}.

\subsection{Iterative Training Set Refinement (ITSR)}
\label{sec:itsr}
In order to improve the robustness against contaminated datasets, we propose an iterative refinement of the training set. This method reduces the influence of likely anomalies for further autoencoder training, thus learning a more accurate model of the normal data distribution. 
If the adversarial autoencoder is trained with an imposed unimodal prior, e.g., a multivariate Gaussian, we expect the normal instances to cluster around the mode of the prior in latent space. This assumption is reasonable whenever instances of the normal class can be expected to be similar, e.g., in quality control. If anomalies are contained in the training set, we observe that the AAE maps them to low-likelihood regions of the prior (see Fig.~\ref{fig:Example}). Anomalies either form their own clusters if they belong to reoccurring patterns (e.g., anomalies from a separate class), or will be represented sparsely and distant from the peak. 
In order to identify likely anomalies, standard outlier detection methods such as 1-class SVM \cite{scholkopf2001estimating} are applied to the representations of training images in the lower-dimensional latent space. The 1-class SVM receives as a hyperparameter an upper bound on the expected fraction of anomalies via the parameter $\nu$. In our experiments, we use a 1-class SVM with RBF kernel and fix $\nu=0.02$, since we assume to have no knowledge of the true anomaly rate.
If available, however, knowledge of the true anomaly rate can be incorporated here.

The output of the 1-class SVM is a decision boundary, and a list of all normal data points. All other data points can be considered potential anomalies, and can be either completely removed from the training set, or weighted to contribute less to the overall loss than normal points.
After modifying the training set the autoencoder is re-trained, yielding representations that better capture the true data manifold of the normal class, and with less incentive to reconstruct outliers accurately.
In the following we describe our proposed training procedure in more detail.

First, every training sample $\mathbf{x}_i$ is associated with a weight $w_i$, which is used to compute a weighted reconstruction loss for the autoencoder:
\begin{equation*}
L_{\mathbf{w}} = \sum_{i=1}^{N} w_i L(\mathbf{x}_i, g(f(\mathbf{x}_i))) ~~~.
\end{equation*}
The autoencoder is trained to minimize the weighted reconstruction loss, where weights can change over time. The same associated sample weight $w_i$ is used in the adversarial training procedure.

To iteratively refine the training set to make the model robust to anomalies present in training, the training procedure is split into three phases: 
\begin{enumerate}
	\item \textbf{Pretraining:} the AAE is initialized by training on the complete training set for a fixed number of epochs where all weights are set to the identical value $w_i=1$.	
	This is the starting point for anomaly detection in latent space with 1-class SVM in the subsequent step.
	\item \textbf{Detection and Refinement:} a 1-class SVM is trained on the latent representations with a constant expected anomaly rate $\nu$, yielding a set of candidate anomalies denoted $\hat{A}$. All instances within $\hat{A}$ are assigned a new weight $w_i = 0$, thereby removing it from further training. The model is then trained on the reduced training set $\mathbf{X} \setminus \hat{A} $ for a short number of epochs. These two steps are repeated $d$ times where each repetition increases the total number of detected training anomalies. By iteratively excluding candidate anomalies, the model of the normal class is refined.
	\item \textbf{Re-training:} 
	after detecting anomalies in the training set and refining the model of the normal class, the model is re-trained such that reconstruction errors on detected anomalies increase. This can be achieved by setting $ w_i < 0, \mathbf{x}_i \in \hat{A} $, forcing a better separability of the two classes. The method, however, created many false positive detections in the previous phase, which with this strict reweighting, would erroneously be forced to be reconstructed worse.
	Since refining the model on normal observations still leads to good reconstructions of those false positive observations (they resemble the true normal observations), we define a threshold $ T_{\mathrm{retrain}} = p_{0.8}(L(\mathbf{x}, f(g(\mathbf{x}))) |\mathbf{x}\in \hat{A}) $ which is used as a new decision threshold for reweighting the potential anomalies, i.e., $ w_i = w_{anomaly} < 0$ if $ L(\mathbf{x}_i, f(g(\mathbf{x}_i))) >  T_{\mathrm{retrain}}$, else $w_i = 0,  \mathbf{x}_i \in \hat{A}$. This forces the model to learn a higher reconstruction error and lower likelihood for the detected candidate anomalies that exceed the threshold $T_{\mathrm{retrain}}$.
\end{enumerate}

\begin{figure*}[ht]
	\centering
	\begin{subfigure}{\columnwidth}
			\resizebox{.9\linewidth}{!}{ 
			\includegraphics[trim=0 10 0 0,clip,width=\columnwidth]{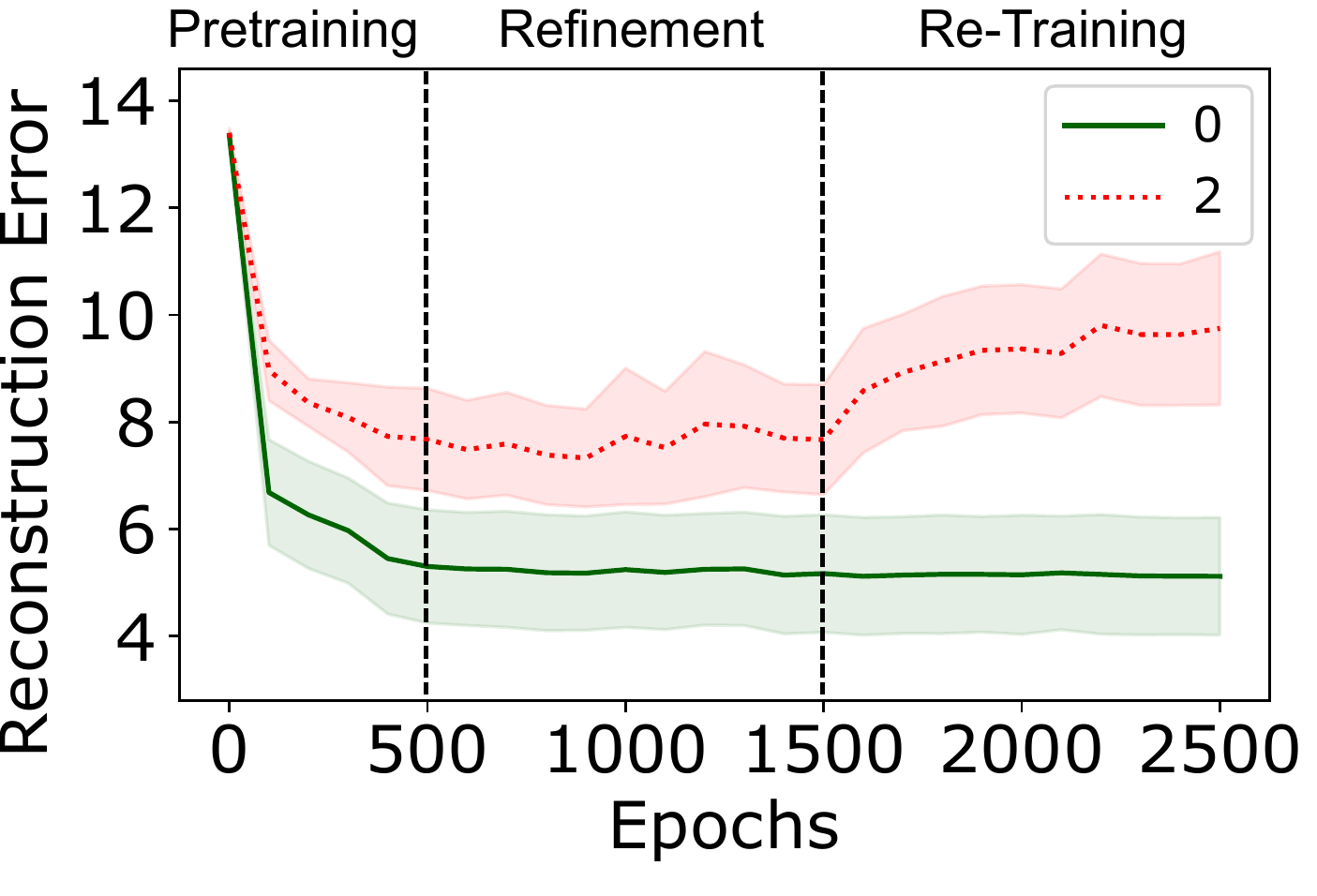} 
			}
			\caption{}
		\end{subfigure} \hfill
		\begin{subfigure}{\columnwidth}
			\resizebox{.9\linewidth}{!}{ 
			\includegraphics[trim=0 5 0 0,clip,width=\columnwidth]{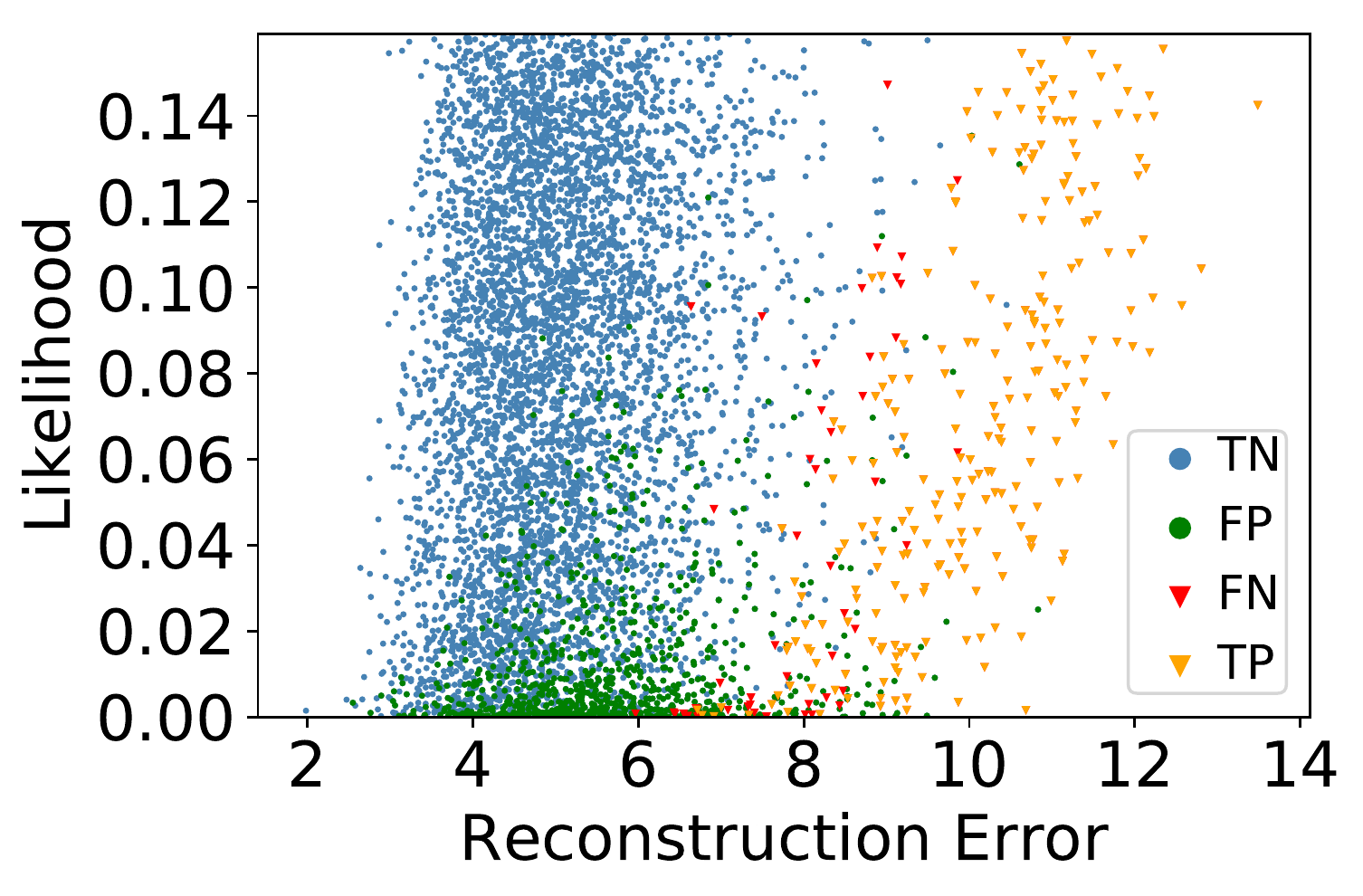} 
			}
			\caption{}
		\end{subfigure} 
	\caption{Increasing the robustness of anomaly detection with iterative training set refinement. 
		(a) Reconstruction error over the three phases of ITSR. We show the mean reconstruction error trained on MNIST, where $95\%$ of the images are from the normal class (digit '0', green, solid line), and $5\%$ are anomalies (digit '2', red, dashed line). The shaded area shows the standard deviation. (b) Reconstruction and likelihood of all points in the training set after ITSR. Colors indicate the final classification result produced by the 1-class SVM in ITSR: true normal (blue), false positive (green), true positive (orange), false negative (red). Iteratively refining and re-training our model increases separability between normal and anomalous observations.
		Additionally, the expected behavior that anomalies that falsely lie in a high-density region are badly reconstructed becomes even more evident.} 
	\label{fig:ocsvm}
\end{figure*}

Our proposed ITSR model yields an autoencoder which over time focuses more and more on the reconstruction of normal images and matching their latent-distribution to the expected prior, thereby increasing the robustness for true normal observations in both training and test set. In Fig.~\ref{fig:ocsvm}, results for applying our ITSR model on MNIST with $5\%$ anomalies in training are presented. While during the refinement phase the model is trained to robustly represent the normal class, the model increases separability between normal and anomalous observations during re-training (Fig.~\ref{fig:ocsvm}(a)). Moreover, the expected effect that anomalies represented in high-likelihood regions have a high reconstruction error becomes more distinct (Fig.~\ref{fig:ocsvm}(b)).
In Section~\ref{sec:experiments}, we also discuss how to set the parameters $\nu$ for detecting candidate anomalies and the threshold for re-training in more detail.   

\section{Experimental Setup and Discussion of Results}
\label{sec:experiments}
\paragraph{Experimental Setup} 
Anomaly detection is evaluated on the classical MNIST \cite{lecun1998gradient} dataset, and the more recent and more complex Fashion-MNIST \cite{xiao2017online} database containing gray-level images of different pieces of clothing such as T-shirts, boots, or pullovers, which in contrast has more ambiguity between classes.
Throughout our experiments, we use the original train-test splits, resulting in $60000$ potential training ($6000$ per class) and $10000$ test observations. From the available classes in both datasets, we define one class to be normal and a second class to be anomalous for training.
In the case of MNIST, we arbitrarily select digit '0' as normal, and digit '2' as anomaly.
For Fashion-MNIST, we conduct two experiments with increasing difficulty: in both cases, the class 'T-shirt' is defined as normal. In the first experiment anomalies are from the class 'Boot', which is easy to distinguish from T-shirts. In the second experiment, anomalies stem from the class 'Pullover', which is visually more similar to T-shirts, except for longer sleeves, and thus harder to detect.  
The final training data consists of the two previously defined classes, but only $\alpha = \{ 5\%, 1\%, 0.1 \% \}$ of the instances are from the anomalous class. 
The experiments with an anomaly rate $\alpha = \{ 1\%, 0.1 \% \}$ show that our approach performs also favorable if anomalies occur even less frequently. 
Since our main focus is to improve AE-based anomaly detection, we thus focus on a comparison to the methods that only partially use the techniques that we combine in our approach.

\paragraph{Architecture}
Encoder and decoder in the conventional autoencoder both consist of $2$ fully-connected layers with $1000$ units each. The ReLU activation function is used in all layers, with the exception of the final encoder layer (using linear activation), and the last decoder layer (using sigmoid activation). The latent space is restricted to $32$ dimensions. This architecture is used in all experiments, but for AAEs the latent space is reduced to $2$ dimensions. On MNIST, training is stopped after $4000$ epochs, on Fashion-MNIST after $10000$ epochs, using Adam \cite{kingma2014adam} to adapt learning rates.
For the AAE, a discriminator network is added, consisting of $2$ fully connected layers with $1000$ units each, and sigmoid activation in the last layer. Following \citet{makhzani2015adversarial}, batch normalization is performed after each layer in the encoder. As latent prior we assume a 2-dimensional Gaussian $p(z) = [ N(0,1) ]^2$. Training is stopped after $1000$ epochs.\\
Our proposed method ITSR is applied to the same AAE architecture. First, pretraining is performed for $500$ epochs, then $d = 10$ repetitions (each $100$ epochs) of the detection and refinement phase with $\nu=0.02$ are computed. Retraining is done for $1000$ epochs on MNIST and $500$ epochs on Fashion-MNIST. \\
For the combined likelihood and reconstruction anomaly score that is used as detection criterion for AAE and ITSR, the $90\%$ percentile $T_{\mathrm{rec}} = p_{0.90}(L(\mathbf{x}, \mathbf{x}^*) |\mathbf{x}\in\mathbf{X})$ of reconstruction errors, and the $10\%$ percentile of likelihoods $T_{\mathrm{prior}} = p_{0.10}(f(\mathbf{x}) | \mathbf{x}\in\mathbf{X})$ are used. Conventional AEs use the same reconstruction-based threshold $T_{\mathrm{rec}}$.

\paragraph{Test set split} 
Our proposed model increases robustness to anomalies that are present during training. In order to evaluate whether this also increases robustness to unobserved types of anomalies, we evaluate on an independent test set, and split the anomalies into classes that were observed during training, and those that were not part of training (e.g. new digit classes in MNIST). For the set containing observed anomalies, the test set contains  all normal test observations and observations from the class of anomalies that was present during training. The set containing unobserved anomalies consists again of the entire set of normal test instances, and all instances from classes that were never observed during training. For example for MNIST, the test set containing observed anomalies consists of images of digit '0' and digit '2' ($1000$ observations each). The set with unobserved anomalies contains again all images of digit '0' and all images of anomaly classes '1', '3'-'9' ($1000$ observations each). This results in a ratio of normal to anomalous observations in the test sets of 1:1 and 1:8, respectively, but does not affect the anomaly rate during training.


\paragraph{Setting of parameters $\nu$ and Re-training Threshold}
For our proposed Iterative Training Set Refinement, the parameters $\nu$, which influences how many candidate anomalies are detected during training, and the threshold for re-training are crucial.\\
In fact, setting the parameters depends on the prior knowledge about the data. If the normal data are expected to be very homogeneous (e.g., in quality inspection), they will lie close in latent space and potential anomalies will most likely lie outside this region, so a smaller $\nu$ will suffice. If, on the other hand, the normal class is very heterogeneous (e.g., if different types of anomalies are expected), more normal observations will spread over latent space and more candidate anomalies (i.e., a larger $\nu$) needs to be detected to ‘catch’ the true anomalies. In practice the true anomaly rate is not known precisely, but our results show that it is not necessary to have a precise estimate for $\nu$ (we know the true anomaly rate in the training data but fix $\nu=0.02$) and that our proposed approach is robust. \\
For the threshold for re-training, the relation between data homogeneity and parameter value is reversed: since the threshold parameter for re-training defines the corresponding percentile of the reconstruction error, a large value is possible for a homogeneous normal class, whereas a lower value is required for heterogeneous normal data.

\begin{figure*}[ht]
	\centering
	\begin{subfigure}{0.3\textwidth}
		\resizebox{.99\linewidth}{!}{ \includegraphics[width=\textwidth]{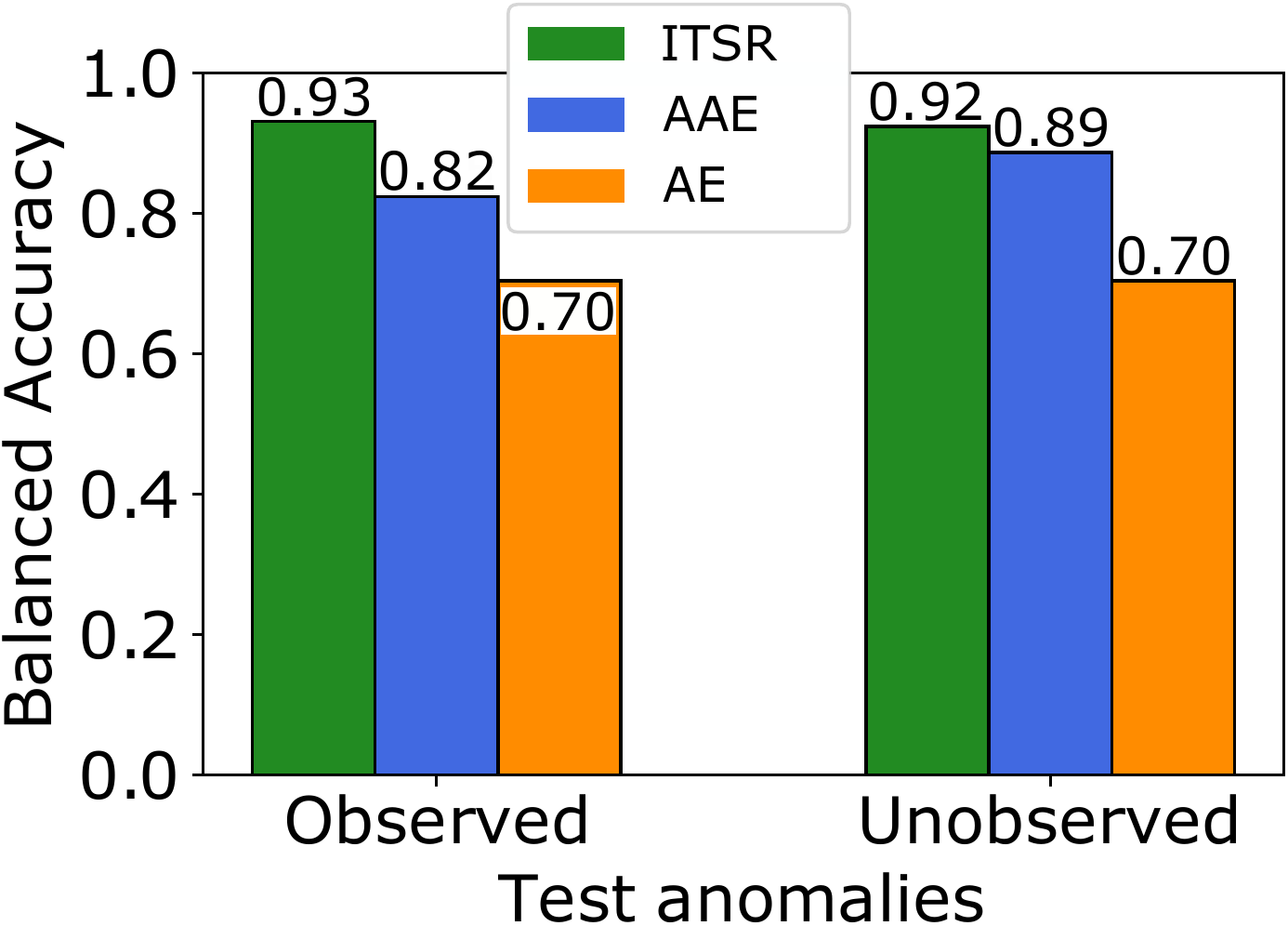} }
		\caption{MNIST: '0' vs. '2'}
	\end{subfigure} \hfill
	\begin{subfigure}{0.3\textwidth}
		\resizebox{.99\linewidth}{!}{ \includegraphics[width=\textwidth]{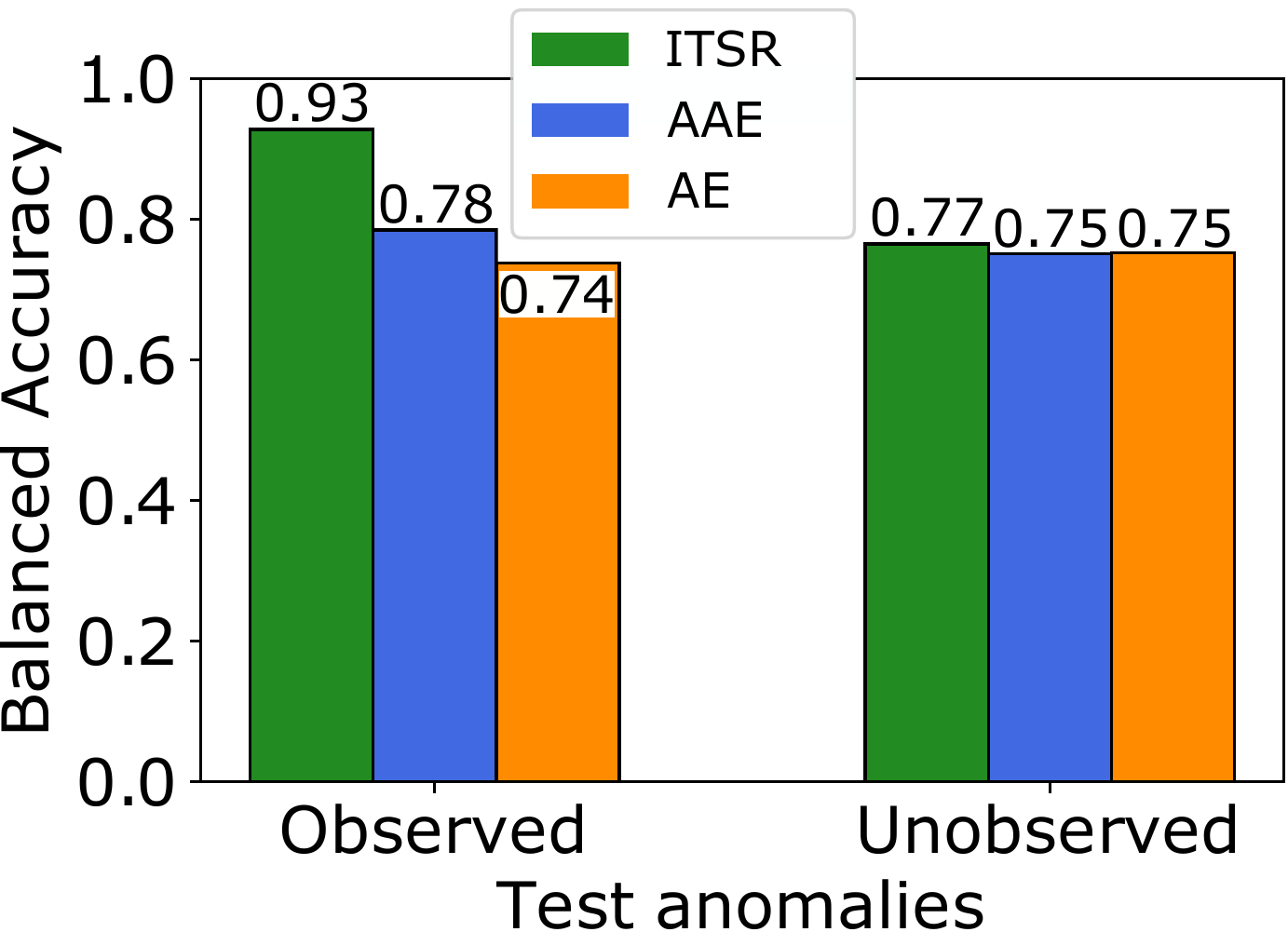} }
		\caption{Fashion-MNIST: T-shirt vs. Boot}
	\end{subfigure} \hfill
	\begin{subfigure}{0.3\textwidth}
		\resizebox{.99\linewidth}{!}{  \includegraphics[width=\textwidth]{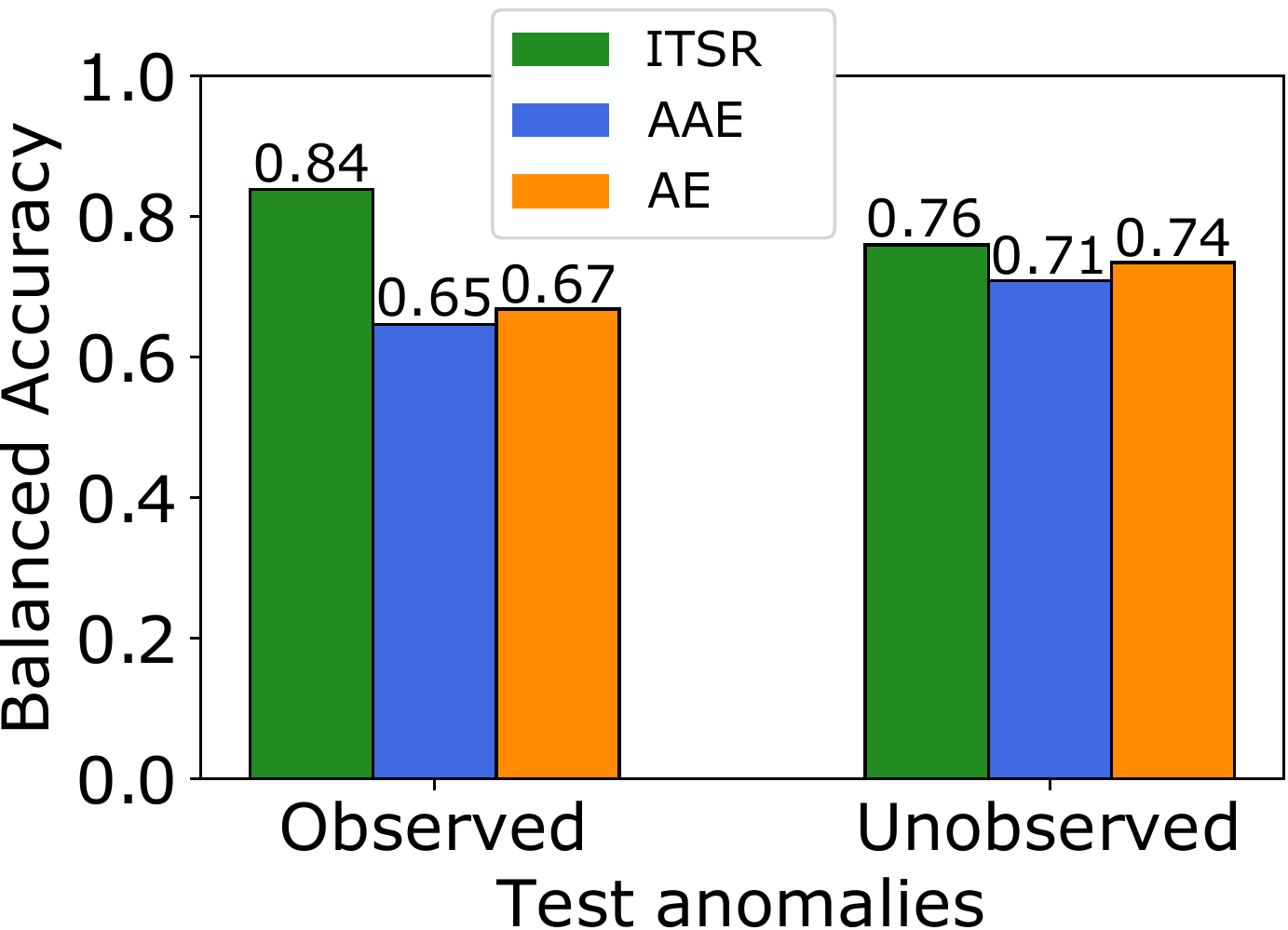} }
		\caption{Fashion-MNIST:~T-shirt~vs.~Pullover}
	\end{subfigure} \hfill
	\par\bigskip
	\begin{subfigure}{0.3\textwidth}
		\resizebox{.99\linewidth}{!}{ \includegraphics[width=\textwidth]{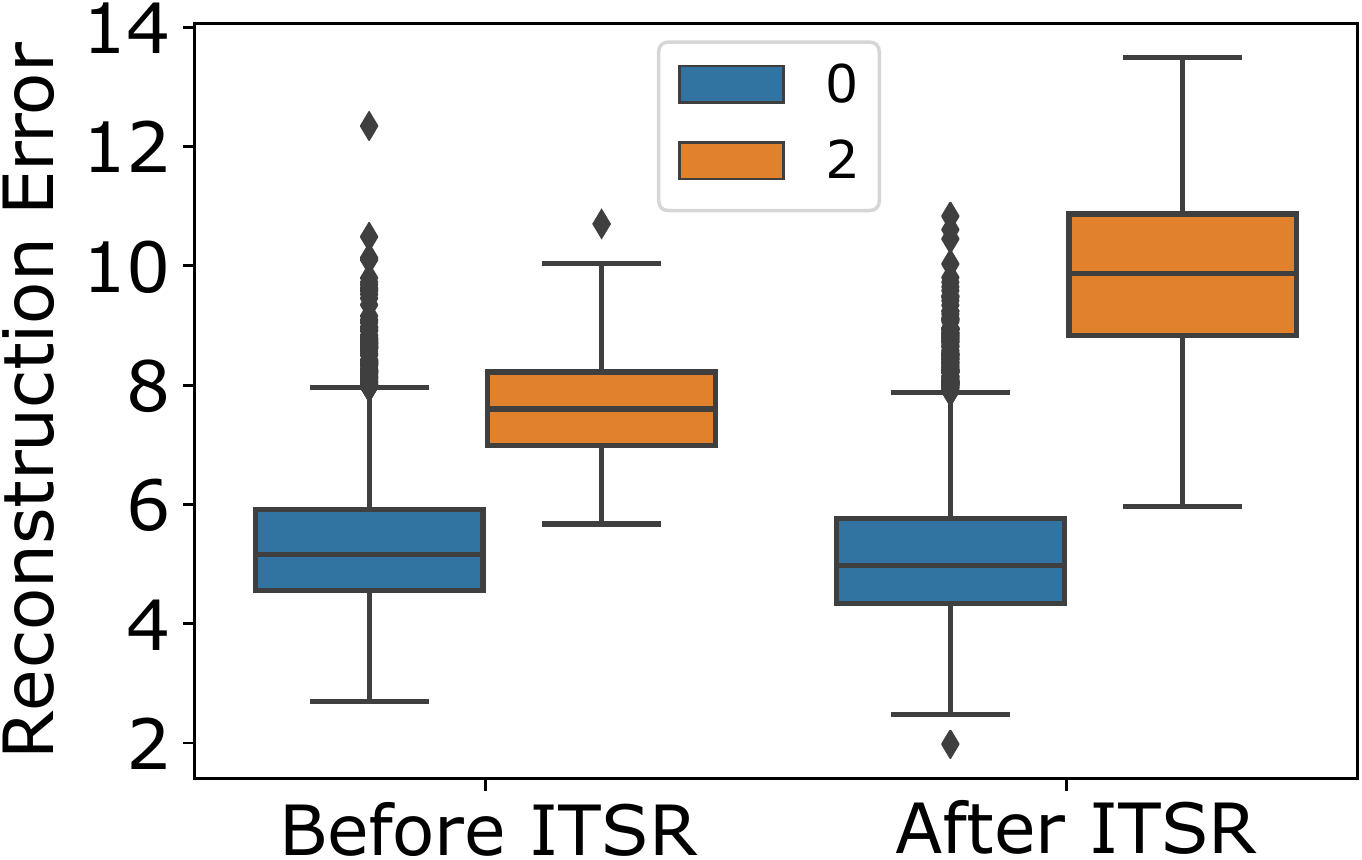} }
		\caption{MNIST: '0' vs. '2'}
	\end{subfigure} \hfill
	\begin{subfigure}{0.3\textwidth}
		\resizebox{.99\linewidth}{!}{ \includegraphics[width=\textwidth]{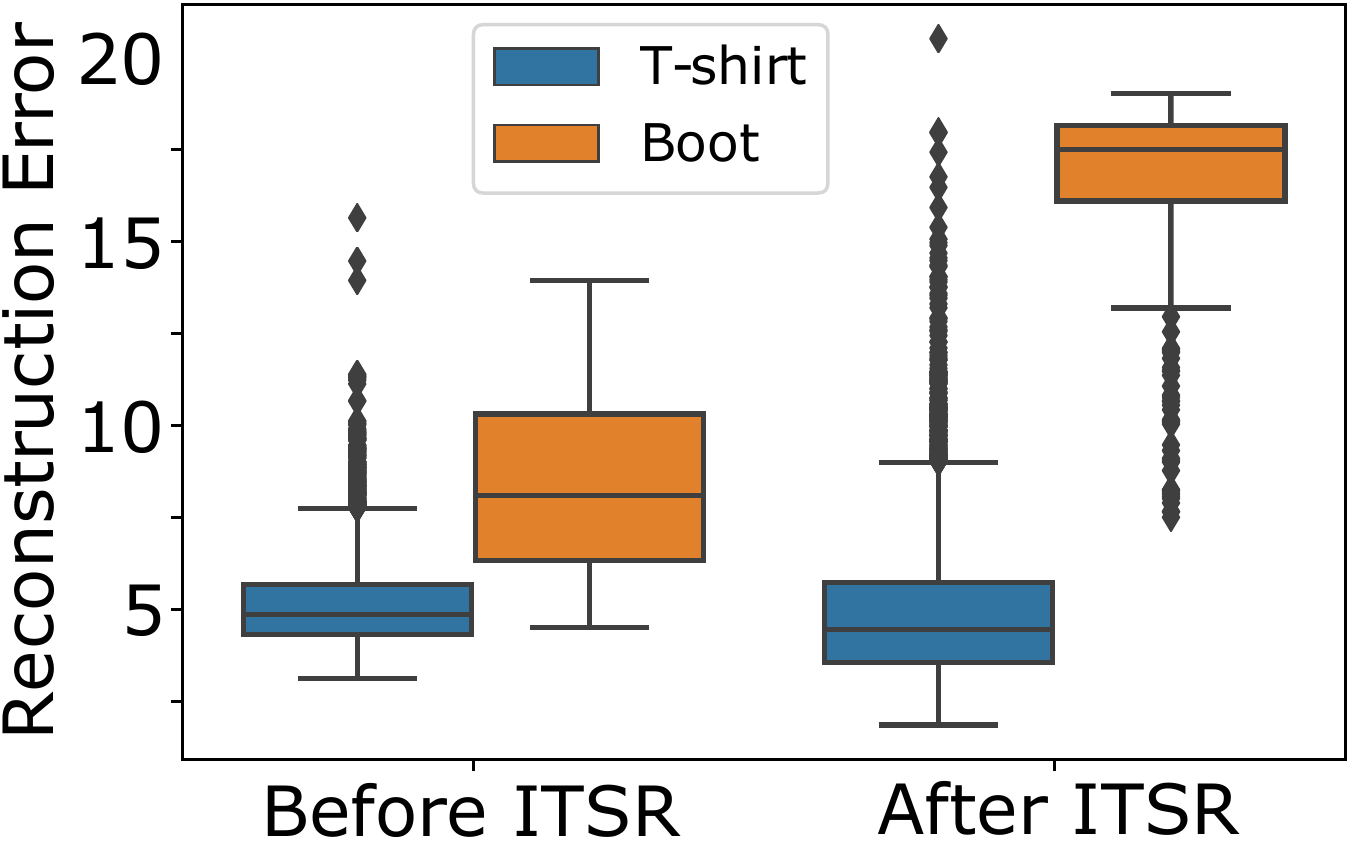} }
		\caption{Fashion-MNIST: T-shirt vs. Boot}
	\end{subfigure} \hfill
	\begin{subfigure}{0.3\textwidth}
		\resizebox{.99\linewidth}{!}{  \includegraphics[width=\textwidth]{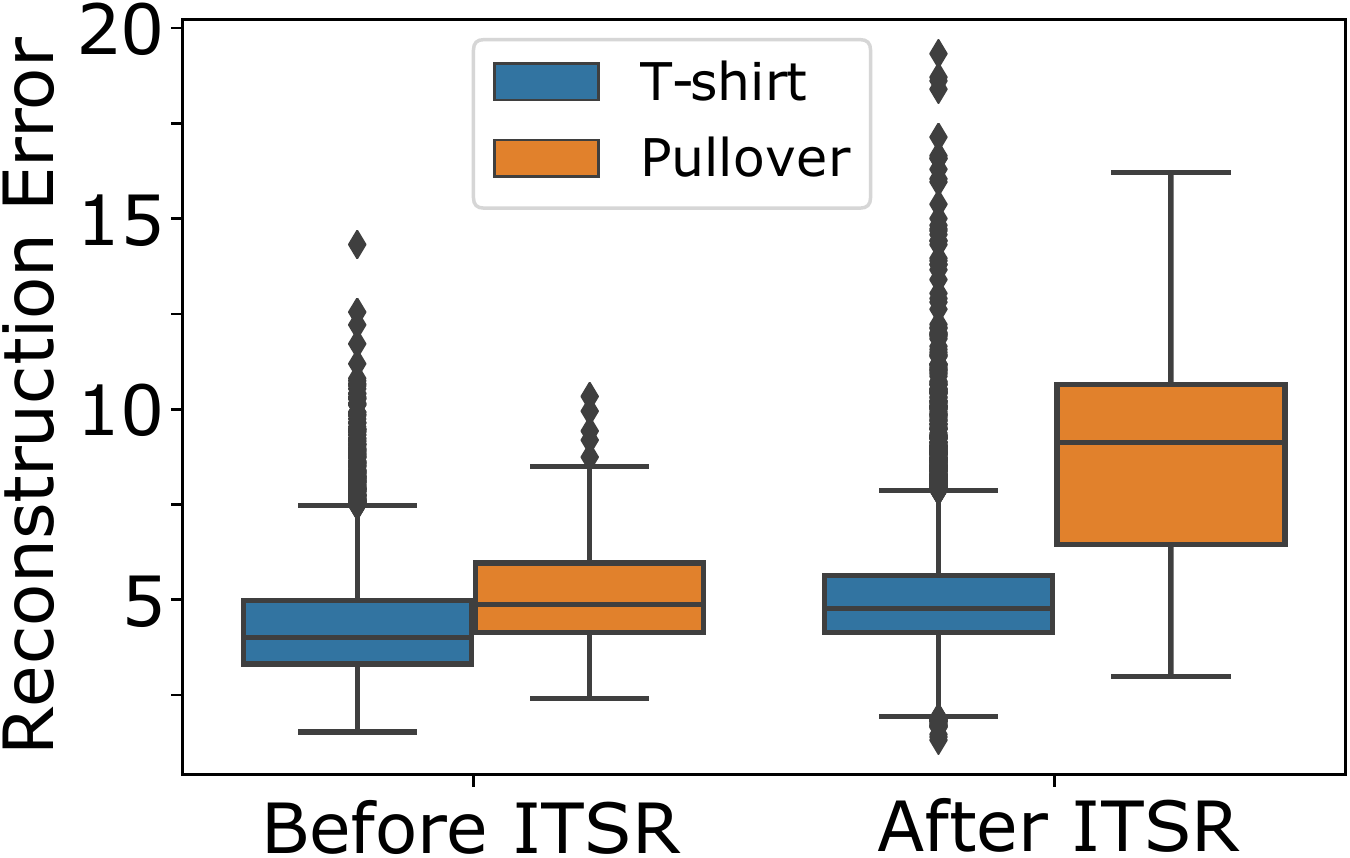} }
		\caption{Fashion-MNIST: T-shirt vs. Pullover}
	\end{subfigure}  
	\caption{Anomaly detection performance on an independent test set with an anomaly rate of $5\%$ in training. The images differ in their experimental setup as follows. (a),(d): MNIST data with digit '0' normal class and digit '2' anomaly class. (b)-(c),(e)-(f): Fashion-MNIST data with class 'T-shirt' defined as normal and 'Boot' ((b),(e)) or 'Pullover' ((c),(f)) as anomalous. \\
	(a)-(c): We compare the BAcc of AE, AAE, and ITSR on a test set containing only types of anomalies observed during training (left), and a set with unobserved anomalies (right). The detection of anomalies during training in ITSR increases the robustness against the type of anomalies contaminating the training set, while the performance on novel anomalies is similar or slightly better. (d)-(f) Reconstruction error for ITSR for normal (blue) and anomalous (orange) training observations.
	The reconstruction error before and after ITSR is shown. Normal images are always accurately reconstructed, but due to ITSR the error for anomalies increases, thus facilitating anomaly detection.}
	\label{fig:results}
\end{figure*}

\paragraph{Results and discussion}
Fig.~\ref{fig:results} shows that for all investigated scenarios with $\alpha=5\%$ anomalies in the training set our ITSR model yields better balanced accuracy than standard autoencoders and adversarial autoencoders. The AAE without refinement improves the anomaly detection performance on MNIST, but has no beneficial effect for Fashion-MNIST. 
The results show the desired increased robustness to the presence of anomalies in the training set, in particular for the observed anomalies that stem from the same class that contaminates the training set, and which pose the greatest challenge for standard AEs.
ITSR improves the balanced accuracy compared to the conventional AE by more than $30\%$  for the experiment on MNIST (Fig.~\ref{fig:results}(a)),
and by more than $20\%$ over the AAE in general. The performance improvement over the AAE is greatest ($30\%$) for the most difficult case of detecting 'Pullover' anomalies in the Fashion-MNIST dataset, with 'T-shirt's being normal (see Fig.~\ref{fig:results}(c)). Addtional experiments in Table~\ref{table:result} show that even with a decreased anomaly rate $\alpha = \{1\%, 0.1\%\}$ our method still performs favorable.

Comparing the performance on anomaly classes that were observed or unobserved during training, we find that standard AEs and AAEs perform similarly on both types. ITSR results in higher accuracy for anomalies observed during training, which is the desired effect. This observations even holds if the training set only contains $0.1\%$ anomalies, or in other words, when the training data is almost completely normal. Furthermore, our model performs at par or slightly better than the other methods on unobserved anomalies. It is expected that the effect for unobserved anomalies is smaller, since they cannot influence training, and any improvement can only come from a more accurate model for the normal class. We thus conclude that iterative refinement of the training set improves anomaly detection with autoencoders in general, without negatively affecting the detection of novel types of anomalies.

\begin{table}[ht]
	\caption{Additional results for anomaly detection on an independent test set with anomaly rate $\alpha = \{1\%, 0.1\%\}$ in training. The defined normal and anomaly class are described in detail in Fig.~\ref{fig:results}(a)-(c). We split evaluation into a test set containing only types of anomalies observed during training (upper), and a set with unobserved anomalies (lower).}
	\centering
	\begin{tabular}{l|lll|lll}
		 &  \multicolumn{3}{c|}{$\alpha = 1\%$} & \multicolumn{3}{c}{$\alpha = 0.1\%$} \\
		 Method &  \rotatebox[origin=l]{90}{MNIST}    & \rotatebox[origin=l]{90}{\parbox[c]{2.5cm}{Fashion-MNIST:\\T-shirt vs. Boot} }   & \rotatebox[origin=l]{90}{\parbox[c]{2.5cm}{Fashion-MNIST:\\T-shirt vs. Pull.} }   & \rotatebox[origin=l]{90}{MNIST}       & \rotatebox[origin=l]{90}{\parbox[c]{2.5cm}{Fashion-MNIST:\\T-shirt vs. Boot}  }     & \rotatebox[origin=l]{90}{\parbox[c]{2.5cm}{Fashion-MNIST:\\T-shirt vs. Pull.} }      \\
		\hline
		\multicolumn{7}{c}{Observed type of anomalies} \\
		\hline
		AE &     0.69     & 0.74        &0.74        &0.68            &   0.74         &   0.73        \\
		AAE &       0.91      &    0.89     & 0.70       &   0.90         &   0.89         &       0.71    \\
		Ours (ITSR) &      0.94      &      0.92   &    0.81    &      0.91      &   0.90         &           0.80\\
		\hline
		\multicolumn{7}{c}{Unobserved type of anomalies} \\
		\hline
		AE &    0.69         & 0.73        &0.74        & 0.68           & 0.73           & 0.73          \\
		AAE &        0.91     &     0.78    &    0.78    &     0.89       &   0.77         &       0.79  \\
		Ours (ITSR) &      0.93  &        0.79 &       0.81 &        0.90    &   0.80         &  0.80
	\end{tabular}
	\label{table:result}
\end{table}

In order to understand the cause for the improved robustness, Fig.~\ref{fig:results}(d)-(f) show the reconstruction errors on training set before and after ITSR, separately for the normal and anomaly classes. We only visualize the case of $\alpha = 5\%$, even though similar observations can be made for decreased anomaly rates.
We observe only minor changes for the normal class, but a strongly increased reconstruction error for anomalies after ITSR. This implies that the ITSR model has learned to robustly represent the normal class in the low-dimensional latent space and reconstruct it to the original space, while becoming insensitive to the anomalies present in training.
There is still some overlap between the reconstruction errors of the two classes, but the increased separation results in a higher balanced accuracy.

\section{Conclusion}
A novel method called ITSR for anomaly detection in images is presented, which exploits the capabilities of adversarial autoencoders in order to address the shortcoming of conventional autoencoders in the presence of anomalies in the training set. Our method compares favorably to state-of-the art methods, and its increased robustness reduces the need for a clean training dataset, and thus the need for expert information. In practice this makes the ITSR method very attractive for scenarios where it is known that the anomaly rate is very low, e.g., in quality inspection. Instead of letting experts inspect a potentially very large training set and picking only normal instances, an unprocessed dataset can be used, leaving it to ITSR to exclude potential anomalies from training. ITSR works directly in the latent space of the AAE, and is a general method to focus the learning process on the true manifold of the normal majority class. No label information is necessary for this approach, but obviously our method can be extended to a semi-supervised setting, or an active learning approach, where an interactive query for labels for instances close to the border identified by the 1-class SVM is performed. Although presented only on image data in this article, our approach easily translates to other high-dimensional data types, e.g., spectrograms or time series.

\bibliography{RobustAnomalyDetection}
\bibliographystyle{icml2019}

\end{document}